\pdfoutput=1

\documentclass[11pt]{article}

\usepackage{acl}

\usepackage{times}
\usepackage{latexsym}

\usepackage{adjustbox}

\usepackage[T1]{fontenc}

\usepackage[utf8]{inputenc}

\usepackage{microtype}

\usepackage{inconsolata}

\newcommand{\plus}{\raisebox{.4\height}{\scalebox{.6}{\plus}}}
\usepackage{xcolor}	
\usepackage{booktabs}
\usepackage{graphicx}
\usepackage{multirow}
\graphicspath{{figures/}}
\usepackage{lipsum}

\usepackage{bbm}
\usepackage{amsmath}
\usepackage{float}
\usepackage{subcaption}
\usepackage{booktabs}
\usepackage{cleveref}

\usepackage{makecell}
\usepackage{mathtools}
\usepackage{amssymb}
\usepackage{siunitx}
\usepackage{etoolbox}           
\renewcommand{\bfseries}{\fontseries{b}\selectfont} 
\robustify\bfseries             
\newrobustcmd{\B}{\bfseries} 
\usepackage{comment}

\usepackage{enumitem}

\title{Decoding Reading Goals from Eye Movements}

\author{Omer Shubi$^{*,1}$, Cfir Avraham Hadar\thanks{Equal contribution.}$\ \ ^{,1}$, Yevgeni Berzak$^{1,2}$ \\
 $^1$Faculty of Data and Decision Sciences, \\
 Technion - Israel Institute of Technology, Haifa, Israel \\
 $^2$Department of Brain and Cognitive Sciences, \\
 Massachusetts Institute of Technology, Cambridge, USA \\
 \texttt{\{shubi,kfir-hadar\}@campus.technion.ac.il},
\texttt{berzak@technion.ac.il} \\
}

\begin{document}

\maketitle





\begin{abstract}
Readers can have different goals with respect to the text that they are reading. Can these goals be decoded from their eye movements over the text? In this work, we examine for the first time whether it is possible to distinguish between two types of common reading goals: \emph{information seeking} and \emph{ordinary reading} for comprehension. Using large-scale eye tracking data, we address this task with a wide range of models that cover different architectural and data representation strategies, and further introduce a new model ensemble. We find that transformer-based models with scanpath representations coupled with language modeling solve it most successfully, and that accurate predictions can be made in \emph{real time}, long before the participant finished reading the text. We further introduce a new method for model performance analysis based on mixed effect modeling. Combining this method with rich textual annotations reveals key properties of textual items and participants that contribute to the difficulty of the task, and improves our understanding of the variability in eye movement patterns across the two reading regimes.\footnote{
Code is available at the following anonymous link: \url{https://anonymous.4open.science/r/Decoding-Reading-Goals-from-Eye-Movements/}. 
}
\end{abstract}

\section{Introduction}

Reading is a ubiquitously practiced skill that is indispensable for successful participation in modern society. When reading, our eyes move over the text in a sequence of \emph{fixations}, during which the gaze is stable, and rapid transitions between fixations called \emph{saccades} \citep{rayner1998eye,schotter2025beginner}. This sequence is generally hypothesized to contain rich information about language comprehension in real time and the nature of the reader's interaction with the text. 

In daily life, a reader may have one or several \emph{goals} that they pursue with respect to the text. For example, they may read the text closely or skim it to obtain the gist of the text's content, they may proofread it, or they may be seeking specific information of interest. Each such goal can impact online linguistic processing and on the corresponding eye movement behavior while reading. Despite the many reading goals readers pursue in everyday life, research on eye movements in cognitive science as well as work that integrated eye movements data in NLP and machine learning have primarily focused on one reading regime, which can be referred to as \emph{ordinary reading}. In this regime, the reader's goal is typically general comprehension of the text. 
Although widely acknowledged, other forms of reading received much less attention and remain understudied \citep{radach2004theoretical}.

In this work, we go beyond ordinary reading and ask whether broad reading goals can be reliably decoded from the pattern of the reader's eye movements over the text. We focus on the distinction between ordinary reading and \emph{information seeking}, a highly common reading regime in everyday life, where the reader is interested in obtaining specific information from the text. 

Decoding reading goals from eye movements has practical implications in several areas. In education, it can enable real-time monitoring of students’ engagement, facilitating targeted interventions to support effective reading and information-seeking strategies. For user-centric applications, it allows dynamic content adaptation, such as highlighting relevant information when users are seeking specific details. In assistive technologies, it can provide real-time support for special populations, such as helping elderly users navigate complex websites by identifying and addressing their information-seeking needs.

Prior work suggests that on average across participants and texts, there are substantial differences in eye movement patterns between information seeking and ordinary reading \citep{hahn2023modeling,malmaud_bridging_2020,shubi2023cogsci}. However, it is currently unknown whether there is sufficient signal in eye movement record for automatic decoding of the reading goal given eye movements of a single participant over a single textual item. Furthermore, little is known about the factors that contribute to the difficulty of this task. 

In this work, we address this gap by conducting a series of experiments on reading goal decoding. 
Our main contributions are the following:
\begin{itemize}
    \item \textbf{Task}: We introduce a new decoding task: given eye movements from a single participant over a passage, predict whether they engaged in ordinary reading for comprehension or in information seeking.
    
    \item \textbf{Modeling}: We adapt and apply to this task 12 different state-of-the-art predictive models for eye movements in reading. We further introduce an ensemble model which leverages the diversity of predictions from single models. 
    
    \item \textbf{Evaluation} We systematically characterize the generalization ability of the models across new textual items and new participants. We find that the models that perform best use scanpath sequence representations as well as the text. We further demonstrate that it is feasible to perform the task \emph{online} and make accurate predictions long before the participant finished reading the text.
    
    \item \textbf{Error Analysis}: We introduce mixed effects modeling of model logits as a general method for analyzing model performance as a function of different properties of the data. Differently from univariate error analyses common in the literature, this method allows examining each feature of interest while controlling for all other features, and taking into account item and subject dependencies in the data. Combining this method with rich data annotations reveals key interpretable axes of variation that contribute to task difficulty, and provides new insights on the data itself. 
\end{itemize}

\begin{figure*}[ht]
    \centering
    \includegraphics[width=1\linewidth]{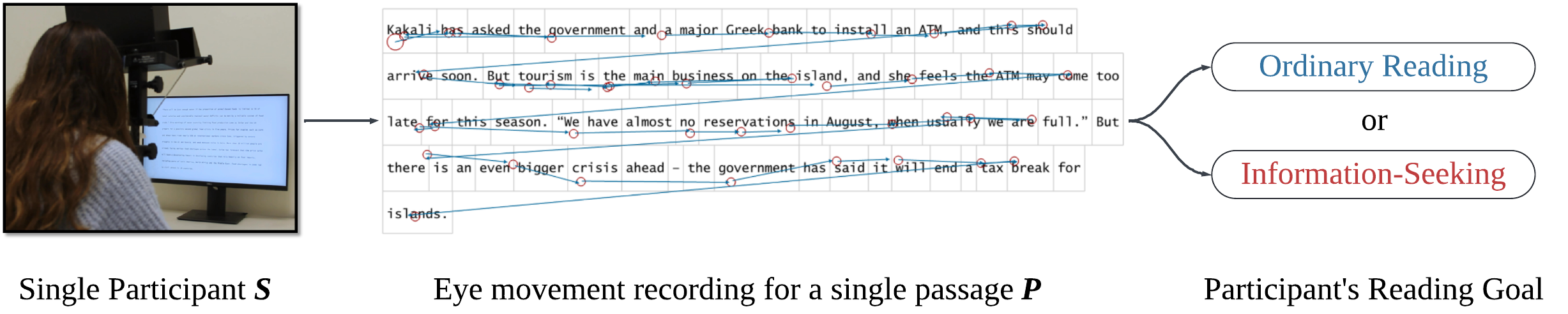}
    \caption{Proposed task: decoding whether a reader is seeking specific information or reading for general comprehension, given their eye movements over a single passage. In the eye movements image, circles represent fixations, and lines represent saccades. Bounding boxes mark word Interest Areas (fixations within the box are assigned to the respective word).
    }
    \label{fig:high_level}
\end{figure*}

\section{Task}

We address the task of predicting whether a reader is engaged in ordinary reading for comprehension or in seeking specific information, based on their eye movements over the text.
Let $S$ be a participant, $P$ a textual passage, and $E_P^S$ the recording of the participant's eye movements over the passage.
Given a ground-truth mapping $\mathcal{C}(S,P) \rightarrow \{\text{Information Seeking},\text{Ordinary Reading}\}$, we aim to approximate $\mathcal{C}$ with a classifier $h$:
\[
h:(E^P_S,P)\rightarrow 
\begin{cases} 
\text{Information Seeking} \\
\text{Ordinary Reading} 
\end{cases}
\]
Where the passage $P$ is an optional input, such that the classifier can be provided only with the eye movement data $E^P_S$ or with both the eye movements and the underlying text. We assume that the participant has not read the paragraph previously.

The information seeking regime is a general framework for addressing goal based reading. It is operationalized by presenting the participant with a question $Q$ prior to reading the passage. This question prompts the participant to seek specific information in the text which results in hundreds of different text-specific tasks. We assume that the classifier does not receive the question nor any information on the participant, which makes the task relevant for real-world scenarios where users are anonymous and no information is available about their specific information seeking goal. \Cref{fig:high_level} presents the task schematically.

\section{Modeling}
\label{sec:models}

Eye movements during reading consist of fixations and saccades \cite{rayner1998eye,schotter2025beginner}, and present a highly challenging case of temporally \textit{and} spatially aligned multimodal data, where fixations are both temporal and correspond to specific words in the text. Recently, a number of general purpose predictive models for eye movements have emerged, each typically evaluated on a different task. 
Here, we adjust and deploy them for a single task, which allows a systematic comparison of architectural and data representation strategies. The models can be broadly divided along three primary axes, the modalities used (eye movements-only, or eye movements and text), how eye movement information is represented (global feature averages across the text, single word, single fixation or an image of the fixation sequence), and for the multimodal approaches, the nature of the text representations and strategy for combining them with eye movements.  

\subsection{Eye Movements-only Models}

These models use only eye movement information, without taking into account the text. Such models are valuable in common scenarios where the underlying text for the eye movement recording is not available. It is also the go-to approach when the eye-tracking calibration is of low quality, leading to imprecise information on the location of fixations with respect to the text. This is a highly common situation, especially with web-based eye-tracking and lower grade eye-tracking devices. Beyond practical considerations, the eye movements-only approach allows assessing the added value of textual information for our task. The models include:

\begin{itemize}[leftmargin=*]
\item \textbf{Logistic Regression} model with 9 global eye movement measures capturing average fixation and saccade metrics. 
\item \textbf{BEyeLSTM - No Text}, similar to BEyeLSTM \citep{reich2022inferring} (see below), but without the text features.
\item \textbf{Vision Models} Following the approach of \citep{bhattacharya2020relevance}, we use two vision models, ViT \citep{dosovitskiy2021an} and ConvNext v2 \citep{woo2023convnext}, that represent the scanpath as an image without the underlying text, where fixations are depicted as circles with diameter proportional to the fixation length. See examples of input images in \Cref{fig:scanpath-as-image} in \Cref{app:sec-models-desc}. 
\end{itemize}

\subsection{Eye Movements and Text Models}

We further adjust a number of recent multimodal models that combine eye movements with textual information. The models encode textual information in two ways. The first is using contextual word embedding representations commonly used in NLP. The second is via linguistic word property features, including word length, word frequency and surprisal \citep{hale2001probabilistic,levy2008expectation}, which are motivated by their ubiquitous effects on reading times \citep[][among others]{rayner2004,kliegl2004,rayner2011}. 

The models implement three primary strategies for combining the two modalities at progressively later stages of processing: (i) in the model input, (ii) merging them within intermediate model representations, or (iii) with architectures that fuse the modalities using cross attention mechanisms after each modality has been processed separately.
Furthermore, since eye movements in reading are both temporally and spatially aligned with the underlying text, the models can be categorized based on how they capture this alignment: (i) by aggregating eye movements for each \emph{word}, thereby focusing on spatial alignment; or (ii) by aggregating eye movement information for each individual \emph{fixation}, which explicitly encodes both spatial and temporal correspondences between eye movements and text. We adjust the following models:

\begin{itemize}[leftmargin=*]
    \item \textbf{RoBERTa-Eye-W} \citep{Shubi2024Finegrained}: A multimodal transformer model that combines word embeddings with word-level eye movement features at the input layer.
    \item \textbf{RoBERTa-Eye-F} \citep{Shubi2024Finegrained}: similar to the above, but with fixation-level representations.
    \item \textbf{MAG-Eye} \citep{Shubi2024Finegrained}: Injects word-level eye movement features into intermediate transformer representations. 
    \item \textbf{PLM-AS} \citep{Yang2023PLMASPL}: Reorders word embeddings based on fixation sequences and processes them with an RNN.
    \item \textbf{Haller RNN} \citep{haller2022eye}: Processes fixation-ordered word embeddings with concatenated eye movement features via an RNN.
    \item \textbf{BEyeLSTM} \citep{reich2022inferring}: Combines fixation sequences and global features with an LSTM and a linear projection layer.
    \item \textbf{Eyettention} \citep{Deng_Eyettention2023}: Aligns word and fixation sequences using cross-attention between a RoBERTa encoder and an LSTM fixation encoder.
    \item \textbf{PostFusion-Eye} \citep{Shubi2024Finegrained}: Combines RoBERTa word representations and convolution-based fixation features using cross-attention and shared latent projection.
\end{itemize}
See \Cref{app:sec-models-desc} for additional details about each of the models, and \Cref{fig:model-comparison-1,fig:model-comparison-2} in \Cref{sec:app-model-diag} for model diagrams.

\subsection{Logistic Ensemble}

As shown in \Cref{tab:main-results} the examined models have diverse predictive behaviors. We therefore introduce a \textbf{Logistic Ensemble}: a 12-feature logistic regression model that predicts the reading goal from the probability outputs of our 12 models. 

\subsection{Baselines}

When examining the utility of eye movements for a prediction task, it is important to benchmark models against simpler approaches that do not require eye movement information \citep{Shubi2024Finegrained}. We therefore introduce the following two baselines:

\begin{itemize}[leftmargin=*]
    \item \textbf{Majority Class} Assigns the label of the majority class in the training set to all the trials in the test set. Since our data is balanced (see below), this baseline is equivalent to random guessing.
    \item \textbf{Reading Time (per word)} Total reading time per word, computed by dividing the total reading time of the paragraph by the number of words in the paragraph. This behavioral baseline does not require eye-tracking and is motivated by the analyses of \citet{hahn2023modeling}, \citet{malmaud_bridging_2020} and \citet{shubi2023cogsci}, which indicate that on average, reading is faster in information seeking compared to ordinary reading.
\end{itemize}

\begin{figure*}[ht]
    \centering
\includegraphics[width=1\textwidth]{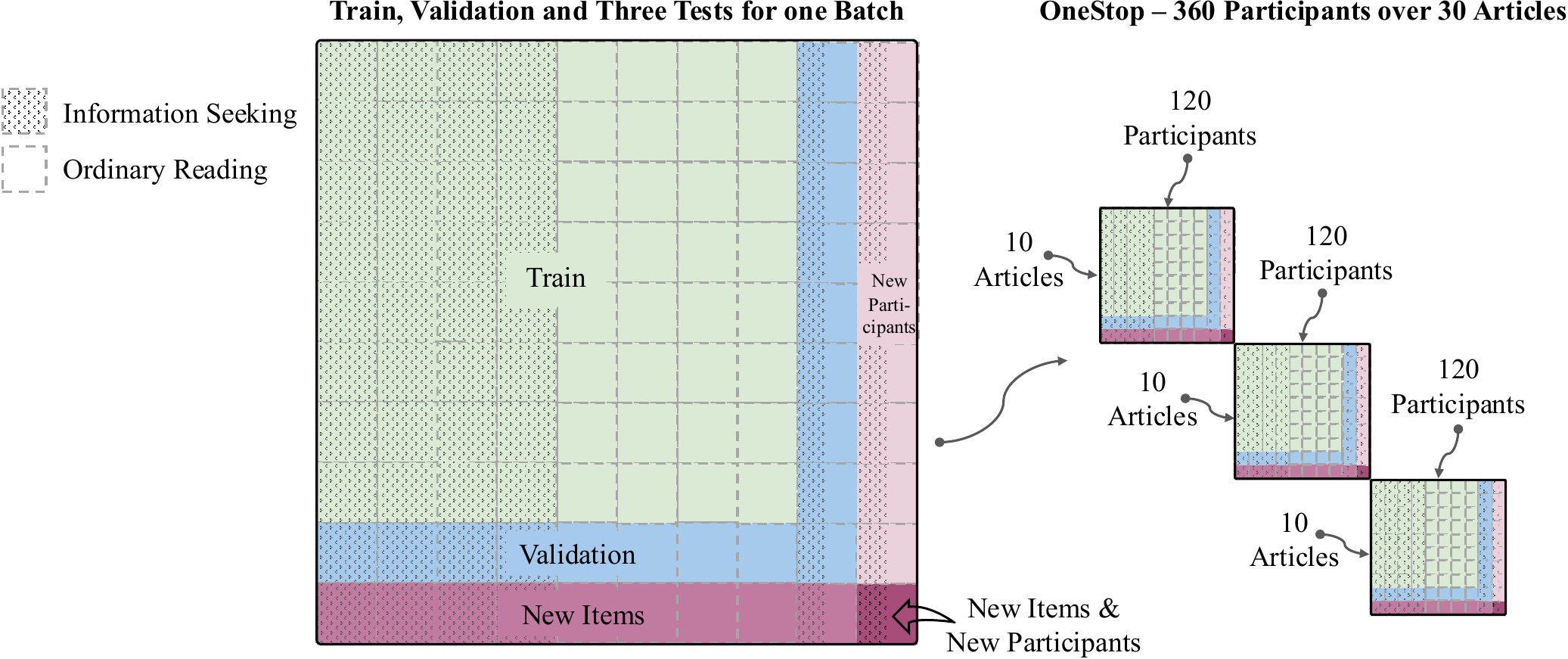}
    \caption{A schematic depiction of one of the 10 splits into train, validation, and the three test sets for one batch of 10 OneStopQA articles and 120 participants. Dashed lines denote information seeking trials. The full data split consists of the union of three such splits. } 
    \label{fig:data-split}
\end{figure*}

\section{Experimental Setup}

\subsection{Data}
\label{subsec:data}

Addressing the proposed task is made possible by OneStop Eye Movements \citep{onestop2025preprint}, the first dataset that contains broad coverage eye-tracking data in both ordinary reading and information seeking regimes. The textual materials of OneStop are taken from OneStopQA \citep{berzak_starc_2020}, a multiple-choice reading comprehension dataset that comprises 30 Guardian articles from the OneStopEnglish corpus \citep{vajjala_onestopenglish_2018}. Each article is available in the original (Advanced) and simplified (Elementary) versions. Each paragraph has three multiple choice reading comprehension questions that can be answered based on any of the two paragraph difficulty level versions. Each question is paired with a manually annotated textual span, called the \emph{critical span}, which contains the vital information for answering the question. An example of a OneStopQA paragraph along with one question and its critical span annotation is provided in \Cref{tab:annotation-example} in \Cref{sec:app-data}.

Eye movements data for OneStopQA were collected in-lab from 360 adult native English speakers using an EyeLink 1000 Plus eye tracker. Each participant read a batch of 10 articles (54 paragraphs) paragraph by paragraph. 
The experiment has two between-subjects reading goal tasks: information seeking and ordinary reading. In the information seeking task, participants were presented with the question (without the answers) prior to reading the paragraph. In the ordinary reading task, they did not receive the question prior to reading the paragraph. In both tasks, after having read the paragraph, participants proceeded to answer the question on a new screen, without the ability to return to the paragraph.

Each participant read 54 paragraphs, all of which were either in information seeking or in ordinary reading. Each paragraph was read by 120 participants: 60 in ordinary reading and 60 in information seeking (split equally between the Advanced and Elementary versions of the paragraph). 
Overall, the data consists of 19,438 trials, where a trial is a recording of eye movements from a single participant over a single paragraph. The data is balanced, with 9,718 trials in ordinary reading and 9,720 in information seeking. \Cref{fig:hg_example} in \Cref{sec:app-data} shows example trials for both reading regimes. Additional data statistics are described in \Cref{sec:app-data}.

\subsection{Model Training and Evaluation Protocol}
\label{subsec:eval-protocol}

 We use 10-fold cross validation, addressing three levels of model generalization:
\begin{itemize}[leftmargin=*]
    \item \textbf{New Item} (paragraph): eye tracking data is available during training for the participant but not for the paragraph. 
    \item \textbf{New Participant}: eye tracking data is available during training for the paragraph, but not for the participant.
    \item \textbf{New Item \& Participant}: No training data for the participant nor the paragraph.
    \item \textbf{All}: aggregated results for the three regimes.
\end{itemize}
 
The New Item regime evaluates performance on unseen paragraphs using eye movement data from other texts, which is a relevant scenario for applications such as e-learning. The New Participant regime tests predictions for unseen individuals on familiar passages, reflecting scenarios such as exams where behavioral data for the given materials exist, but not from the tested participant. The New Item \& New Participant evaluation, which addresses zero-shot prediction for an arbitrary unseen reader on an arbitrary unseen passage, is the most challenging and flexible regime.

The data is split into train, validation, and the three test sets separately for each batch of 10 articles and the 120 participants who read the batch. The three batch splits are then combined to form the full split of the dataset. Paragraphs are allocated to the train, validation, and test portions of each batch split at the \emph{article level}, such that all the paragraphs of each article appear in the same portion of the split. This ensures that items in the test set are unrelated in content to items in training and validation. 


Each data split contains 64\% of the trials in the training set, 17\% in the validation set and 19\% in the test sets (9\% New Item, 9\% New Participant and 1\% New Item \& Participant). Aggregated across the 10 splits, 90\% of the trials in the dataset appear in each of the New Participant and New Item evaluation regimes, and 10\% in the New Item \& Participant regime. \Cref{fig:data-split} presents this breakdown for one batch split. 

\textbf{Model Hyperparameters} We perform hyperparameter optimization and model selection separately for each split. We assume that at test time, the evaluation regime of the trial is \emph{unknown}. Model selection is therefore based on the entire validation set of the split. 
Further details regarding the training procedure, including
the full hyperparameter search space for all the
models are provided in \Cref{sec:app-model-training}.

\textbf{Statistical Testing} The samples in the OneStop dataset are not i.i.d; each item is read by multiple participants, and each participant reads multiple items. To account for these dependencies when comparing model performance, we fit linear mixed-effects models with maximal random effects for items and participants \citep{barr_random_2013} using the MixedModels package in Julia \citep{bates_juliastatsmixedmodelsjl_2024}. 

\begin{table*}[ht]
\centering
\resizebox{\textwidth}{!}{%
\begin{tabular}{@{}lccllll@{}}
\toprule
\textbf{Model} & {\makecell{\textbf{Gaze} \\ \textbf{Representation}}} & {\makecell{\textbf{Text} \\ \textbf{Representation}}} & {\makecell{\textbf{New} \\ \textbf{Item}}} & {\makecell{\textbf{New} \\ \textbf{Participant}}} & {\makecell{\textbf{New Item}\\ \& \textbf{Participant}}} & {\makecell{\textbf{All}}} \\
\midrule
Majority Class / Chance & -- & -- 
& ${50.0 _{\pm 0.0}}_{\text{\scriptsize +++}}$  
& ${50.0 _{\pm 0.0}}_{\text{\scriptsize +++}}$ 
& ${50.0 _{\pm 0.0}}_{\text{\scriptsize +++}}$ 
& ${50.0 _{\pm 0.0}}_{\text{\scriptsize +++}}$  \\

Reading Time & -- & -- 
& ${59.0 _{\pm 0.4}}_{\text{\scriptsize +++}}$ 
& ${58.9 _{\pm 1.0}}_{\text{\scriptsize +++}}$ 
& ${60.4 _{\pm 1.2}}_{\text{\scriptsize +++}}$ 
& ${59.0 _{\pm 0.5}}_{\text{\scriptsize +++}}$ \\ \addlinespace[1ex]
\hline
\addlinespace[1ex]
Log. Regression & Global & -- 
& ${62.4 _{\pm 0.3}}_{\text{\scriptsize +++}}^{**}$ 
& ${60.6 _{\pm 1.4}}_{\text{\scriptsize +++}}$ 
& ${{60.8 _{\pm 1.6}}_{\text{\scriptsize +++}}}$ 
& ${61.5 _{\pm 0.8}}_{\text{\scriptsize +++}}^{*}$ \\

BEyeLSTM \citep{reich2022inferring} No Text & Fixations & -- 
& ${71.5 _{\pm 0.6}}_{\text{\scriptsize +++}}^{***}$ 
& ${61.0 _{\pm 1.1}}_{\text{\scriptsize +++}}$ 
& ${61.5 _{\pm 1.5}}_{\text{\scriptsize +++}}$ 
& ${65.9 _{\pm 0.4}}_{\text{\scriptsize +++}}^{***}$ \\

ConvNext v2 & Scanpath Image& -- 
& ${70.4_{\pm 0.5}}_{\text{\scriptsize +++}}^{***}$ 
& ${63.7_{\pm 0.8}}_{\text{\scriptsize +++}}^{**}$ 
& ${64.0_{\pm 0.7}}_{\text{\scriptsize +++}}$ 
& ${66.9_{\pm 0.3}}_{\text{\scriptsize +++}}^{***}$ \\

ViT & Scanpath Image& -- 
& ${70.6_{\pm 0.5}}_{\text{\scriptsize +++}}^{***}$ 
& ${64.4_{\pm 0.8}}_{\text{\scriptsize +++}}^{***}$ 
& ${64.4_{\pm 1.5}}_{\text{\scriptsize +++}}^{*}$ 
& ${67.3_{\pm 0.4}}_{\text{\scriptsize +++}}^{***}$ \\

\addlinespace[1ex]
\hline
\addlinespace[1ex]
RoBERTa-Eye-W \citep{Shubi2024Finegrained} & Words & Emb+LF
& ${64.6_{\pm 0.7}}_{\text{\scriptsize +++}}^{***}$ 
& ${62.5_{\pm 1.3}}_{\text{\scriptsize +++}}^{*}$ 
& ${62.0_{\pm 1.3}}_{\text{\scriptsize +++}}$ 
& ${63.5_{\pm 0.9}}_{\text{\scriptsize +++}}^{**}$ \\

MAG-Eye \citep{Shubi2024Finegrained} & Words & Emb+LF 
& ${52.1 _{\pm 0.3}}_{\text{\scriptsize +++}}$ 
& ${52.3 _{\pm 0.4}}_{\text{\scriptsize +++}}$ 
& ${51.5 _{\pm 0.4}}_{\text{\scriptsize +++}}$ 
& ${52.1 _{\pm 0.2}}_{\text{\scriptsize +++}}$ \\
\addlinespace[1ex]
\hline
\addlinespace[1ex]
PLM-AS \citep{Yang2023PLMASPL} & Fixations Order & Emb 
& ${58.6 _{\pm 0.4}}_{\text{\scriptsize +++}}$ 
& ${59.5 _{\pm 0.5}}_{\text{\scriptsize +++}}$ 
& ${57.5 _{\pm 0.9}}_{\text{\scriptsize +++}}$ 
& ${59.0 _{\pm 0.4}}_{\text{\scriptsize +++}}$ \\

Haller RNN \citep{haller2022eye} & Fixations & Emb 
& ${61.7_{\pm 0.6}}_{\text{\scriptsize +++}}^{*}$ 
& ${61.2_{\pm 1.1}}_{\text{\scriptsize +++}}$ 
& ${60.8_{\pm 1.5}}_{\text{\scriptsize +++}}$ 
& ${61.3_{\pm 0.5}}_{\text{\scriptsize +++}}$ \\

BEyeLSTM \citep{reich2022inferring} & Fixations & LF 
& ${71.4 _{\pm 0.9}}_{\text{\scriptsize +++}}^{***}$ 
& ${61.6 _{\pm 1.1}}_{\text{\scriptsize +++}}$ 
& ${62.2 _{\pm 1.3}}_{\text{\scriptsize +++}}$ 
& ${66.2 _{\pm 0.7}}_{\text{\scriptsize +++}}^{***}$ \\

Eyettention \citep{Deng_Eyettention2023} & Fixations & Emb+LF 
& ${55.8 _{\pm 0.8}}_{\text{\scriptsize +++}}$ 
& ${55.7 _{\pm 1.1}}_{\text{\scriptsize +++}}$ 
& ${55.4 _{\pm 1.8}}_{\text{\scriptsize +++}}$ 
& ${55.8 _{\pm 0.9}}_{\text{\scriptsize +++}}$ \\

PostFusion-Eye \citep{Shubi2024Finegrained} & Fixations & Emb+LF 
& ${88.5_{\pm 0.7}}_{\text{\scriptsize +++}}^{***}$ 
& ${90.3_{\pm 0.6}}^{***}$ 
& ${86.0_{\pm 1.1}}_{\text{\scriptsize +}}^{***}$ 
& ${89.3_{\pm 0.4}}_{\text{\scriptsize +++}}^{***}$ \\

RoBERTa-Eye-F \citep{Shubi2024Finegrained} & Fixations & Emb+LF 
& ${\mathbf{89.9}_{\pm 0.6}}^{***}$ 
& ${\mathbf{90.9}_{\pm 0.4}}^{***}$ 
& ${\mathbf{88.2}_{\pm 0.8}}^{***}$ 
& ${\mathbf{90.3}_{\pm 0.3}}^{***}$ \\

\addlinespace[1ex]
\hline
\addlinespace[1ex]
Logistic Ensemble &  & 
& ${91.3_{\pm 1.7}}^{***}_{\&\&\&}$ 
& ${91.6_{\pm 1.6}}^{***}_{\&}$
& ${88.0_{\pm 3.1}}^{***}$ 
& ${91.3_{\pm 1.2}}^{***}_{\&\&\&}$ 
\\
\bottomrule
\end{tabular}%
}
\small
\caption{Test accuracy results aggregated across 10 cross-validation splits, with 95\% confidence intervals. `Emb' stands for word embeddings, `LF' for linguistic word features such as word length, frequency and surprisal, and `Fix' for fixations. Model performance is compared to the Reading Time baseline using a linear mixed effects model. In R notation: $is\_correct \sim model + (model \mid participant) + (model \mid paragraph)$. Significant gains over this baseline are marked with '*' $p < 0.05$, '**' $p < 0.01$ and '***' $p < 0.001$ in superscript, and significant drops compared to the best model are marked in subscript with '+'. The best performing single model is marked in bold. Significant improvements of the Logistic Ensemble over this model are marked with the subscript '\&'.}
\label{tab:main-results}
\end{table*}

\section{Results}
\label{sec:results}

Test set accuracy results are presented in \Cref{tab:main-results}. In line with prior observations of faster reading in information seeking compared to ordinary reading \citep{hahn2023modeling,malmaud_bridging_2020,shubi2023cogsci}, the Reading Time baseline yields above chance accuracies ($p<0.01$ in all  regimes), thus providing a strong benchmark for the evaluation of eye tracking-based models. Among the 12 examined models, \mbox{RoBERTa-Eye-F} achieves the highest accuracy in all the evaluation regimes. PostFusion-Eye comes second, well ahead of the remaining 10 models. 
The top performing models suggest that a combination of three elements is key for our task: a transformer-based architecture, fixation-level encoding of eye movements, and explicit modeling of the text.

The 10 weaker models tend to perform better on the New Participant regime compared to the New Item regime. Due to the between-subjects design, where all the training and test examples for a given participant have the same label, this could reflect, at least in part, an ability of models to learn participant-specific reading behavior without explicit information on the participant (i.e. identify the participant), which is not directly pertinent to the task at hand. The current experimental setup makes it challenging to adjudicate between these two possibilities. In either case, it is highly non-trivial that models are able to generalize from prior participant data to new items. 

Finally, we find that the Logistic Ensemble improves over the accuracy of the best performing single model RoBERTa-Eye-F in all the regimes, with statistically significant improvements in all but the New Item and Participant evaluation. 
These performance improvements suggest that the information encoded by the different models is to some extent complementary, and that they likely capture different aspects of the eye movement data and the task. Further evidence for that can be obtained by examining the agreement between the models. \Cref{app-fig:model_correlation_by_regime} in \Cref{app-sec:model_correlation_by_regime} depicts the pairwise Cohen's Kappa \citep{cohen1960coefficient} agreement rates across models in the validation data, where we observe mostly moderate agreement rates.  
      

In \Cref{app-sec:additional-res} \Cref{app-fig:roc_curves}, we present the Receiver Operating Characteristic (ROC) curves across the ten cross-validation splits and their corresponding Area Under the ROC Curve (AUROC) scores 
\citep{bradley1997use}. Validation set accuracies are reported in \Cref{tab:eval-results}. The outcomes of these evaluations are consistent with the test results in \Cref{tab:main-results}.


\section{How Quickly Can We Make Accurate Predictions?}

Thus far we examined predictions from a complete recording of eye movements for a paragraph. Can we make accurate predictions before the participant finishes reading the paragraph? \Cref{tab:percent-exp} presents RoBERTa-Eye-F All accuracy given the first 1\%, 5\%, 10\%, 25\% and 50\% of the fixation sequence. While as can be expected, data quantity does impact performance, relatively high accuracy predictions can be obtained even with only the initial 5\% of the fixations, which on average corresponds to the first 1.5 seconds of the eye movements recording. This is an important outcome, which demonstrates the feasibility of performing our task \emph{online}, long before the participant finishes reading a passage.

\begin{table}[ht]
    \centering
    \setlength{\tabcolsep}{3pt} 
    \resizebox{\columnwidth}{!}{%
    \begin{tabular}{@{}lcccccc@{}}
        \toprule
        First \% of Fixations & 1\% & 5\% & 10\% & 25\% & 50\% & 100\% \\
        Average Time (sec) & 0.5 & 1.5 & 2.7 & 6.3 & 12.4 & 24.3 \\
        \midrule
        Accuracy (All) & $61.0_{\pm 3.6}$ & $77.6_{\pm 0.3}$ & $78.9_{\pm 0.4}$ & $82.3_{\pm 2.0}$ & $84.9_{\pm 2.4}$ & $90.3_{\pm 0.3}$ \\
        \bottomrule
    \end{tabular}%
    }
    \caption{RoBERTa-Eye-Fixations accuracy with 95\% confidence intervals as a function of the \% of scanpath data used from the beginning of the paragraph reading.}
    \label{tab:percent-exp}
\end{table}

\section{What Makes the Task Easy or Hard?}

Having established that the prediction task at hand can be performed with a considerable degree of success, we now leverage the best performing single model RoBERT-Eye-F to obtain insights about the task itself. To this end, we introduce a new method for analysis of model performance that uses mixed effects modeling of model logits from data features. This method enables examining which trial features, which were not given to the model explicitly, affect the ability of the model to classify trials correctly. Differently from univariate methods often used for model performance analyses, our approach allows measuring the contribution of each feature \emph{above and beyond} all the other features, while also taking into account the \emph{non-i.i.d} nature of the data, where multiple participants read the same paragraph and multiple paragraphs are read by the same participant. The analysis takes advantage of the rich structure and auxiliary annotations of the OneStop dataset.

We define 10 features that capture various aspects of the trial. These include the following \textbf{participant} features over the item: Reading time before, within, and after the critical span, Paragraph position in the experiment (1-54), and whether after having read the paragraph, the participant answered the given reading comprehension question correctly. We further include the following \textbf{item} (paragraph and question), reader-independent features: Paragraph length (in words), Paragraph difficulty level (Advanced / Elementary), Critical span start location (relative position, normalized by paragraph length), Critical span length (normalized by paragraph length), and Question difficulty (percentage of participants who
answered the question incorrectly). Further details about these features are presented in \Cref{sec:app-features}.

\begin{figure}[ht]
    \centering
    \includegraphics[width=\linewidth]{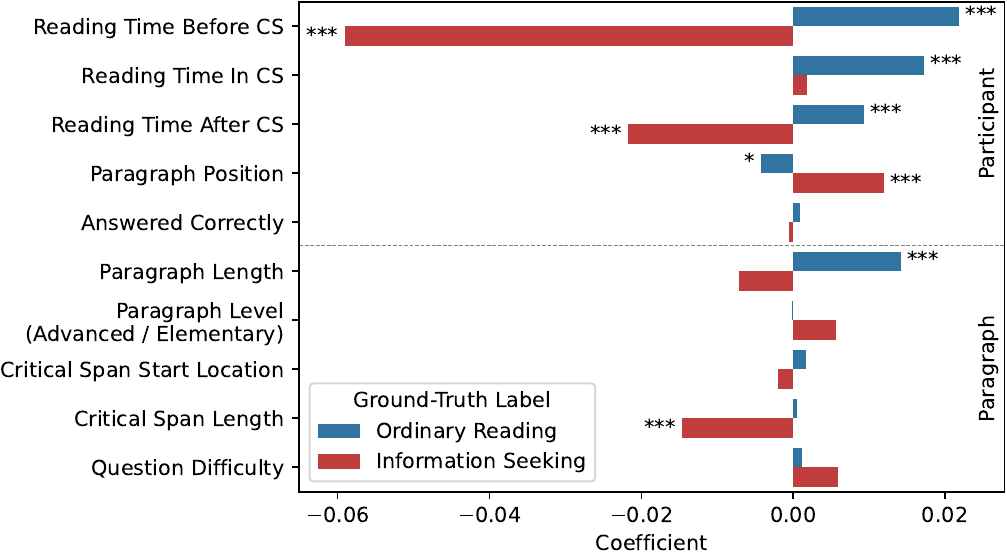}
    \caption{Coefficients from a mixed-effects model that predicts whether RoBERTa-Eye-F's prediction for a given trial is correct from properties of the trial. CS stands for the critical span, the portion of the paragraph that contains the information that is essential for answering the question correctly. Two models are fitted separately for ordinary reading and information seeking trials. Predictors are z-normalized. Depicted are the coefficients of the fitted models after a 10x Bonferroni correction, to mitigate the risk of false positives when testing multiple hypotheses simultaneously. `*' $p<0.05$, `**' $p<0.01$, `***' $p<0.001$.}
    \label{fig:metadata-coeffs}
\end{figure}

To establish the relation of these features to task difficulty, we use a linear mixed effect model that uses these features to predict the probability that the model assigns to the correct label. In R notation:
\begin{equation*}
\begin{array}{l}
P(\text{correct}) \sim \text{feat}_1 + \cdots + \text{feat}_{10} + (1 \mid \text{item}) \\
\quad + (1 \mid \text{participant}) + (1 \mid \text{evaluation regime})
\end{array}
\end{equation*}
where the random effects account for correlations in predictions within participants, items and evaluation regimes\footnote{Random effects structure is simplified not to include slopes due to model convergence issues.}. 
We fit this model separately on the information seeking and ordinary reading trials. To make the contributions of the features to prediction accuracy comparable, we normalize each feature to be a z-score (zero mean and unit variance). We then examine feature contribution via statistical significance, magnitude and sign of the corresponding coefficient. A significant coefficient for a feature indicates that it correlates with task difficulty, the absolute value determines its importance relative to other features, and the sign indicates the direction of the association.

The resulting feature coefficients are presented in \Cref{fig:metadata-coeffs}. In line with the findings of \citet{shubi2023cogsci} on  differences in reading speed between information seeking and ordinary reading around the critical span, we observe that prominent features for correctly classifying both information seeking and ordinary reading trials are reading times before and after the critical span. Faster readers before and after the critical span are easier to correctly classify as information seeking and harder to correctly classify as ordinary reading. Additionally, although not reflected in the reading speed analysis of \citet{shubi2023cogsci}, slower reading within the critical span is also beneficial for correct classification of ordinary reading trials.  Longer paragraphs are also beneficial for classification of ordinary reading. We further find that shorter critical spans facilitate correct classification of information seeking trials, presumably by making information seeking more targeted and the identification of task critical information easier. Paragraph position is also significant in information seeking, suggesting that readers develop more efficient goal oriented reading strategies as they progress through the experiment. Overall, this analysis provides a highly interpretable characterization of both task difficulty and the underlying reading behavior in both reading regimes.

\section{Related Work}

The vast majority of the literature on the psychology of reading and its interfaces with computational modeling is concerned with ordinary reading. However, several studies did address goal-oriented (also referred to as task-based) reading. Most prior work focused on a small number of canonical tasks: skimming, speed reading and proofreading. Several studies found different eye movement patterns in these tasks as compared to ordinary reading \citep{just1982speed,kaakinen_task_2010,schotter2014task,strukelj2018one,chen2023characteristics}. \citet{rayner1996freq} examined differences between ordinary reading and searching through the text for a target word. Prior work also analyzed eye movements during human linguistic annotation, often used for generating training data for NLP tools, such as annotation of named entities \citep{tomanek-2010-cognitive,tokunaga-2017-eye}. 
Differences in reading patterns were further found when readers were asked to take different perspectives on a given text \citep{kaakinen2002perspective} or given different sets of learning goals \citep{rothkopf1979goal}. 

Our work is closest to \citet{hahn2023modeling}, \citet{malmaud_bridging_2020} and \citet{shubi2023cogsci} who analyzed eye movement differences between ordinary reading and information seeking, where the information seeking goal is formulated using a reading comprehension question. All three studies found substantial differences in fixation and saccade patterns in information seeking as compared to ordinary reading, in particular before, within and after the text portions that are critical for the information seeking task. Here, we build on these findings, and examine whether these differences can be leveraged to automatically distinguish between these two reading regimes.

While the above studies focus primarily on descriptive data analysis, \citet{hollenstein2023zuco} took a predictive approach and attempted to automatically classify the reading task from eye movement features. In this study, 18 participants read single sentences from the ZuCo corpus \citep{zuco2020}, and engaged either in ordinary reading or in an annotation of the presence of one of seven semantic relations in the sentence. While this benchmark is conceptually related to the current work, 
it is limited by the nature of the tasks,  
which focus on highly specialized linguistic annotations that are not performed by readers in everyday life. 
In the current study we take a different and more general stance on task based reading, with unrestricted questions that are more representative of the tasks commonly pursued by readers. 
More broadly, our work contributes to a nascent line of  work which uses eye movements in reading for predicting properties of the reader's cognitive state with respect to the text, such as reading comprehension \citep{reich_inferring_2022,Shubi2024Finegrained}, as well as properties of the text itself, including document type \citep{kunze2013know} and readability level \citep{gonzalez-garduno-sogaard-2017-using}.

\section{Summary and Discussion}

Is it possible to decode reader goals from eye movements? We address this question by examining the possibility of automatically differentiating between ordinary reading and information seeking at the challenging granularity level of a single paragraph. We find that it is indeed possible to perform this task with considerable success, even before the participant finished reading, above and beyond reading time. Model comparison reveals that the architecture, the granularity level of the eye movement representation and the inclusion of the underlying text are all important for the task. Our error analysis method leverages the models to further reveal new insights on the factors that determine task difficulty. 


\section{Limitations}
Our study has a number of limitations. The information seeking tasks are over individual paragraphs that span 3-10 lines of text. This leaves out shorter texts (e.g. single sentences) as well as longer texts. It is also restricted to newswire texts, and does not include texts from other genres. Finally, new datasets for the information seeking task, other types of tasks, additional populations (e.g. second language readers, younger and older participants), and datasets in languages other than English are all needed in order to study goal decoding more broadly.

While the current work takes a first step in addressing the proposed task, ample room for performance improvements remains for future work, especially in the two regimes involving unseen participants. New strategies for modeling eye movements with text are likely needed to fully exploit the potential of eye movements for this task. Furthermore, the addressed task is fundamentally limited in that it does not distinguish between different information seeking tasks. A natural direction for future work could address decoding of the specific question that was presented to the participant in the information seeking regime.

\section{Ethics Statement}

This work uses eye movement data collected from human participants. The data was collected by \citet{onestop2025preprint} under an institutional IRB protocol. All the participants provided written consent prior to participating in the eye tracking study. The data is anonymized. Analyses and modeling of eye movements in information seeking are among the main use cases for which the data was collected.

It has previously been shown that eye movements can be used for user identification \citep[e.g.][]{bednarik2005eye,jager2020deep}. We do not perform user identification in this study, and emphasize the importance of not storing information that could enable participant identification in future applications of goal decoding. 
We further stress that future systems that automatically infer reader goals are to be used only with explicit consent from potential users to have their eye movements collected and analyzed for this purpose.






\bibliography{references}

\clearpage
\appendix
\onecolumn
\section*{Appendix}

\section{Models}
\label{app:sec-models}

\subsection{Model Descriptions}
\label{app:sec-models-desc}

\subsubsection*{Global Representation}

\textbf{Logistic Regression} A logistic regression model with global eye movement measures from \citet{meziere2023using}. 
The measures include averages of word reading times, single fixation duration, forward saccade length, the rate of regressions (saccades that go backwards), and skips (words that were not fixated) during first pass reading (i.e. before proceeding to the right of the word). All the features are standard measures from the psycholinguistic literature.

\subsubsection*{Word-based Representations}

\textbf{RoBERTa-Eye-W(ords)} \citep{Shubi2024Finegrained} is a RoBERTa transformer model \citep{liu_roberta_2019} augmented with eye movements. 
This model concatenates word embeddings and word-level eye movement features in the model input.

\textbf{MAG-Eye} \citep{Shubi2024Finegrained} 
 Integrates word-level eye movement features into a transformer-based language model by injecting them into intermediate word representations using a Multimodal Adaptation Gate (MAG) architecture \citep{rahman_integrating_2020}. The text is aligned with eye movements by duplicating each word-level eye movement feature for every sub-word token.

\subsubsection*{Fixation-based Representations}

\textbf{PLM-AS} \citep{Yang2023PLMASPL} 
This model represents the eye movements sequence by reordering contextual word embeddings according to the order of the fixations over the text. This reordered sequence is processed through a Recurrent Neural Network (RNN), whose final hidden layer is used for classification.

\textbf{Haller RNN} \citep{haller2022eye}  
This model is similar to PLM-AS in that it receives word embeddings in the order of the fixations. Differently from PLM-AS, each word embedding is further concatenated with eye movement features.

\textbf{RoBERTa-Eye-F(ixations)} \citep{Shubi2024Finegrained} uses the same architecture as RoBERTa-Eye-W, but represents fixations rather than words. Each fixation input consists of a concatenation of the word embedding and eye movement features associated with the fixation.

{\textbf{BEyeLSTM}} \citep{reich2022inferring} represents both the fixation sequence and textual features, combining LSTMs \citep{hochreiter_long_1997} and global features through a linear layer. 

\textbf{BEyeLSTM - No Text} is a model that processes raw fixation data using an LSTM. The final hidden state of the LSTM is combined with global eye movement features to perform classification. 
The model is inspired by BEyeLSTM \citep{reich2022inferring}, using the same eye movements feature set, without the text representations.

\textbf{Eyettention} \citep{Deng_Eyettention2023} 
is a model that consists of a RoBERTa word sequence encoder and an LSTM-based fixation sequence encoder. It uses a cross-attention mechanism to align the input sequences. We use the adaptation of this model by \cite{Shubi2024Finegrained} for binary classification. 

\textbf{PostFusion-Eye} \citep{Shubi2024Finegrained} 
 is a model that consists of a RoBERTa word sequence encoder and a 1-D convolution-based fixation sequence encoder. It then uses cross-attention to query the word representations using the eye-movement representations, followed by concatenation and projection into a shared latent space. 

\subsubsection*{Image Representations}
\begin{figure}
    \centering
    \includegraphics[width=0.5\linewidth]{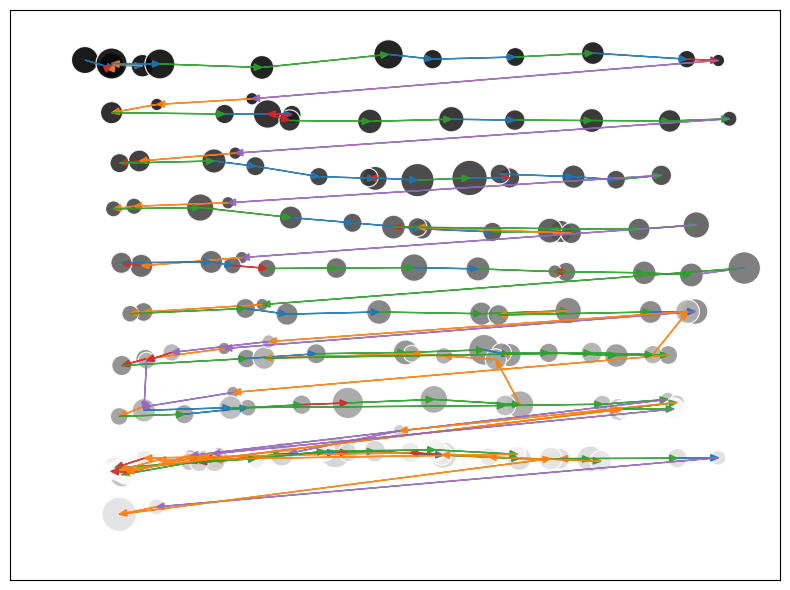}
    \caption{An example of a scanpath as an image as used for the image classification models.}
    \label{fig:scanpath-as-image}
\end{figure}

We represent scanpaths as two-dimensional images, as illustrated in \Cref{fig:scanpath-as-image}. In this visualization, fixations are depicted as circles positioned at their original x-y coordinates from the screen display, with the entire representation cropped to maintain consistent dimensions. The diameter of each circle corresponds to the duration of the fixation. To indicate the sequential progression of the eye movements, we employ a gradient shading scheme. Additionally, we differentiate between saccade types by color-coding them according to the five categories established by \citet{schotter2025beginner} - forward saccade, skip, refixation, return sweep, regression, and another for any saccade typethat does not fall into this categorization. Note that for these features knowledge about the \textit{existance} of text is needed, but not the textual content itself. We use the \textit{convnextv2\_base.fcmae\_ft\_in22k\_in1k
} and \textit{vit\_base\_patch14\_dinov2} versions of the ConvNextv2 and ViT models respectively.

\subsection{Model Diagrams}
\label{sec:app-model-diag}

\begin{figure}[ht]
    \centering
    \begin{subfigure}[t]{0.48\textwidth}
        \centering
        \includegraphics[height=7cm]{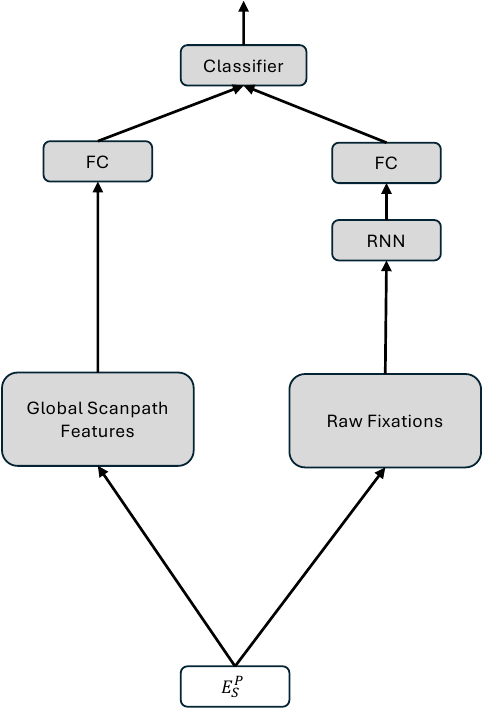}
        \subcaption{BEyeLSTM - No Text}
        \label{fig:beyelstm-notext}
    \end{subfigure}
    \hfill
    \begin{subfigure}[t]{0.48\textwidth}
        \centering
        \includegraphics[height=7cm]{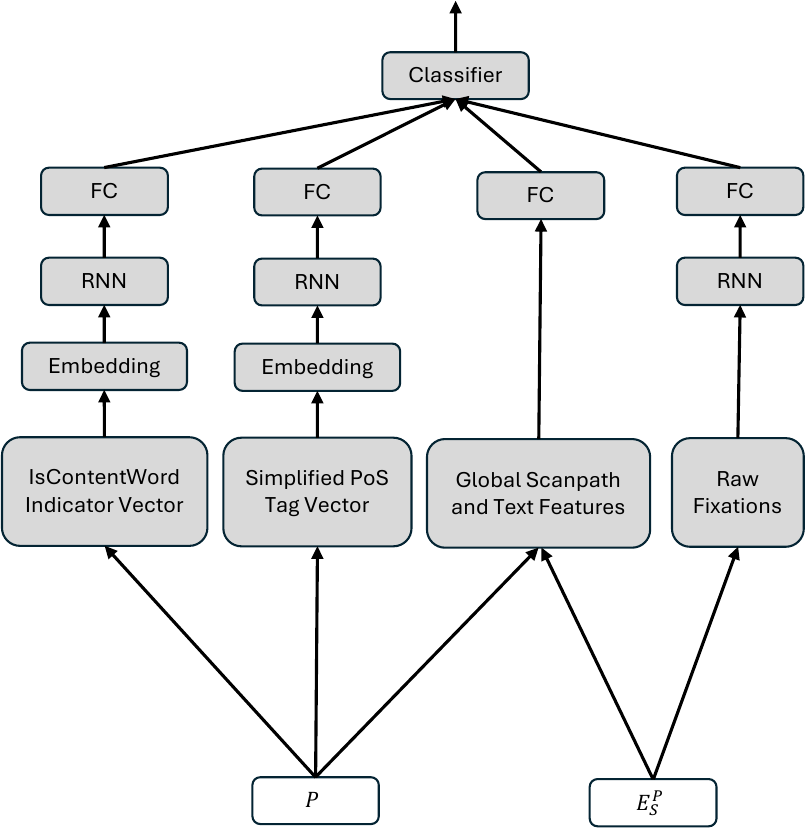}
        \subcaption{BEyeLSTM}
        \label{fig:beyelstm}
    \end{subfigure}
    
    \vspace{1cm}
    
    \begin{subfigure}[t]{0.48\textwidth}
        \centering
        \includegraphics[height=9cm]{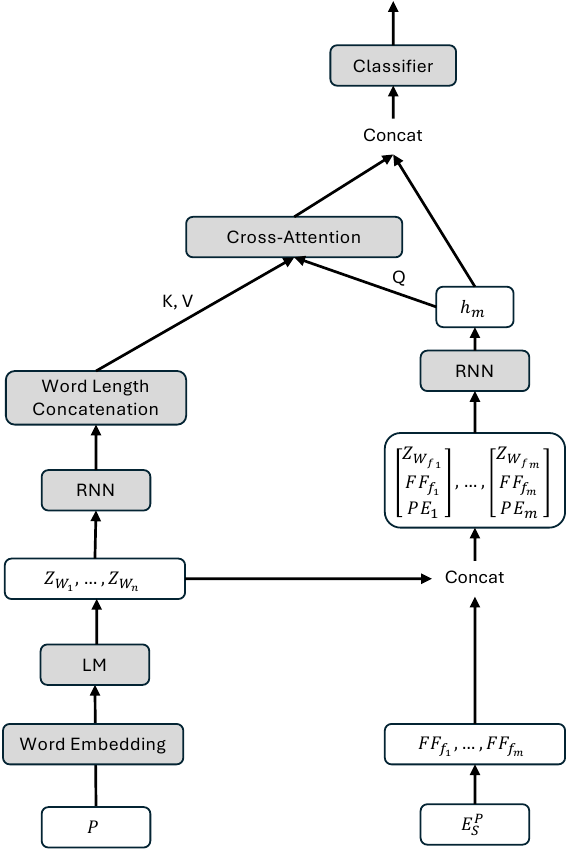}
        \subcaption{Eyettention}
        \label{fig:eyettention}
    \end{subfigure}
    \hfill
    \begin{subfigure}[t]{0.48\textwidth}
        \centering
        \includegraphics[height=9cm]{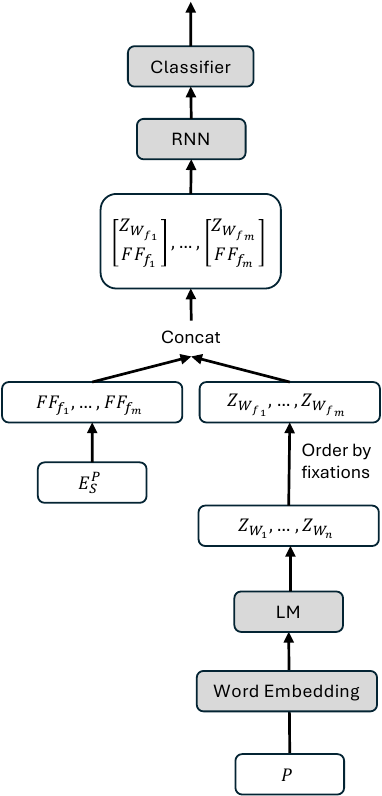}
        \subcaption{Haller RNN}
        \label{fig:haller-rnn}
    \end{subfigure}
    
    \caption{Visualization of the different model architectures (Part 1). $P$ represents the paragraph, $E^P_S$ the eye movements of participant $S$ on $P$. $LM$ stands for a language model, and $FC$ for fully connected layers. $FF_{f_i}$ stands for the fixation features and $w_{f_i}$ for the word corresponding to the $i$-th fixation respectively.}
    \label{fig:model-comparison-1}
\end{figure}

\begin{figure}[ht]
    \centering
    \begin{subfigure}[b]{0.48\textwidth}
        \centering
        \includegraphics[height=9cm]{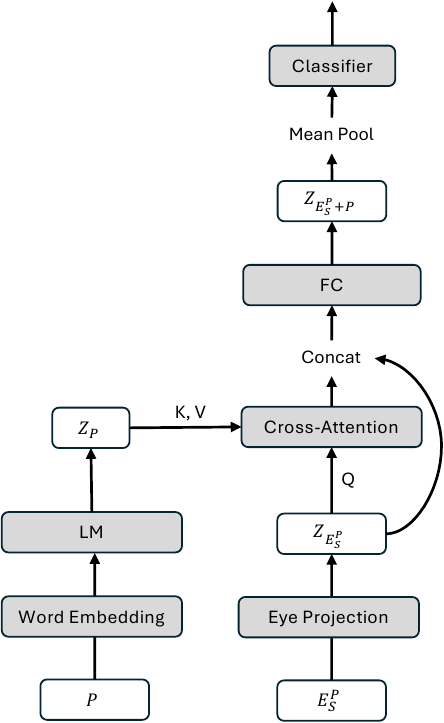}
        \subcaption{PostFusion-QEye}
        \label{fig:postfusion-qeye}
    \end{subfigure}
    \hfill
    \begin{subfigure}[b]{0.48\textwidth}
        \centering
        \includegraphics[height=9cm]{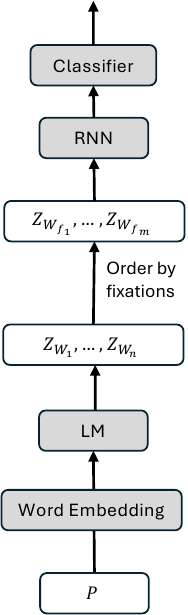}
        \subcaption{PLM-AS}
        \label{fig:plm-as}
    \end{subfigure}
    
    \vspace{1cm}
    
    \begin{subfigure}[b]{0.48\textwidth}
        \centering
        \includegraphics[height=6cm]{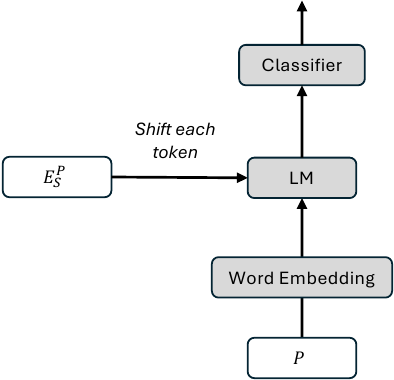}
        \subcaption{MAG-QEye}
        \label{fig:mag-qeye}
    \end{subfigure}
    \hfill
    \begin{subfigure}[b]{0.48\textwidth}
        \centering
        \includegraphics[height=6cm]{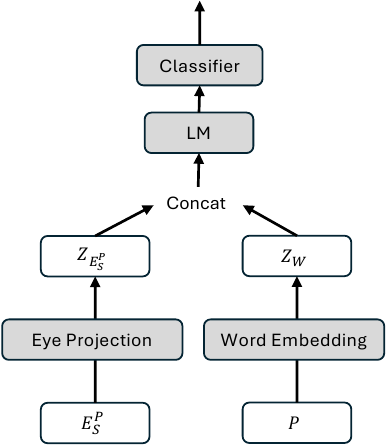}
        \subcaption{RoBERTa-QEye}
        \label{fig:roberta-qeye}
    \end{subfigure}
    
    \caption{Visualization of the different model architectures (Part 2). $P$ represents the paragraph, $E^P_S$ the eye movements of participant $S$ on $P$. $LM$ stands for a language model, and $FC$ for fully connected layers. $FF_{f_i}$ stands for the fixation features and $w_{f_i}$ for the word corresponding to the $i$-th fixation respectively.}
    \label{fig:model-comparison-2}
\end{figure}

\clearpage

\section{OneStop Eye Movements Dataset - Additional Details}
\label{sec:app-data}

The textual data of OneStop consists of $162$ paragraphs, $486$ questions, and $972$ unique paragraph--level--question triplets. The mean paragraph length is 109 words (min: 50; max: 165; std: 28). The mean length of Elementary paragraphs is 97 words (37 before the critical span, 30 inside it, and 30 after it), and of Advanced paragraphs 120 words (48 before the critical span, 34 inside it, and 38 after it). Each question has 20 responses, 10 for the Advanced version and 10 for the Elementary version. The mean experiment duration is approximately one hour. The raw millisecond gaze location is pre-processed into fixations and saccades using the SR Data Viewer software (\texttt{v4.3.210}).

\begin{table*}[ht!]
\small
\caption{An example of a OneStopQA paragraph (Advanced and Elementary version) along with one of its three questions. The critical span is marked in bold red. Adapted from \citet{berzak_starc_2020}.}
\label{tab:annotation-example}
\begin{tabular}{p{0.1\linewidth}|p{0.85\linewidth}}
\hline
    \bf Advanced &  A major international disagreement with wide-ranging implications for global drugs policy has erupted over the right of Bolivia's indigenous Indian tribes to chew coca leaves, the principal ingredient in cocaine. \textbf{\textcolor{red}{Bolivia has obtained a special exemption from the 1961 Single Convention on Narcotic Drugs, the framework that governs international drugs policy, allowing its indigenous people to chew the leaves.}} Bolivia had argued that the convention was in opposition to its new constitution, adopted in 2009, which obliges it to ``protect native and ancestral coca as cultural patrimony'' and maintains that coca ``in its natural state ... is not a dangerous narcotic.'' \\ \hline
    \bf Elementary & A big international disagreement has started over the right of Bolivia's indigenous Indian tribes to chew coca leaves, the main ingredient in cocaine. This could have a significant effect on global drugs policy. \textbf{\textcolor{red}{Bolivia has received a special exemption from the 1961 Convention on Drugs, the agreement that controls international drugs policy. The exemption allows Bolivia's indigenous people to chew the leaves.}} Bolivia said that the convention was against its new constitution, adopted in 2009, which says it must ``protect native and ancestral coca'' as part of its cultural heritage and says that coca ``in its natural state ... is not a dangerous drug.'' \\ \hline
\bf Question   &\textbf{What was the purpose of the 1961 Convention on Drugs?} \\ \hline
\bf Answers &A Regulating international policy on drugs \\
        &B Discussing whether indigenous people in Bolivia should be allowed to chew coca leaves \\
        &C Discussing the legal status of Bolivia's constitution  \\
        &D Negotiating extradition agreements for drug traffickers \\
\end{tabular}
\end{table*}

\begin{figure}[ht!]
    \centering
    \includegraphics[width=1\linewidth]{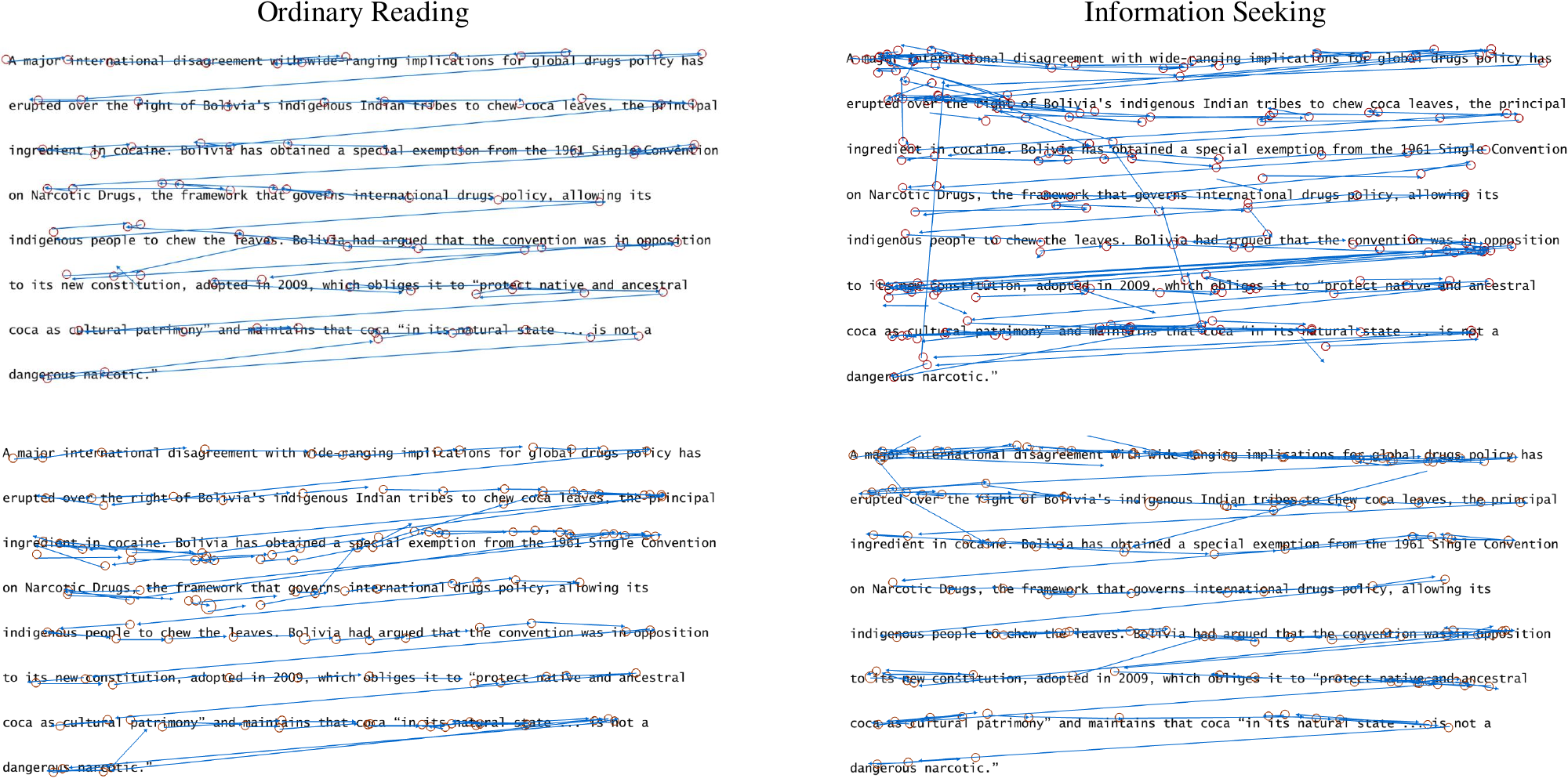}
    \caption{Examples of eye movements over a single passage; left: ordinary reading, right: information seeking. Circles represent fixations, and lines represent saccades. 
    }
    \label{fig:hg_example}
\end{figure}

\clearpage

\section{Model Training and Hyperparameters}
\label{sec:app-model-training}

All neural network-based models were trained using the PyTorch Lighting \citep{falcon_2024} library on NVIDIA A100-40GB and L40S-48GB GPUs.

Since the models we use were developed for different tasks and datasets, we conducted a hyperparameter search for each model. The search space for each model is described below. In all cases, it includes the optimal parameters reported in the work that introduced the model, extended to provide a fair comparison between models.

For all neural models we train with learning rates of $\{0.00001, 0.00003, 0.0001\}$ and dropout of $\{0.1, 0.3, 0.5\}$ following \citet{Shubi2024Finegrained}. Additionally, for all models that make use of word embeddings, we include both frozen and unfrozen language model variants in the search space.

\begin{itemize}
    \item For \textbf{Logistic Regression}, the search space for the regularization parameter C is $\{0.1, 5, 10, 50, 100\}$, with and without an L2 penalty.

    \item Following \citep{reich2022inferring}, for \textbf{BEyeLSTM} and \textbf{BEyeLSTM - No Text}, the search space consists of learning rates $\{0.001, 0.003, 0.01\}$, embedding dimensions $\{4,8\}$ and hidden dimensions $\{64,128\}$.

    \item For \textbf{MAG-Eye} the search space for injection layer index is: $\{0, 11, 23\}$.

    \item Following \citet{Yang2023PLMASPL}, we train \textbf{PLM-AS} and \textbf{Haller RNN}  with dropout rate search space of 0.1, and for PLM-AS, we use LSTM layer counts of ${1, 2}$. Additionally, as in \citep{haller2022eye}, we search over LSTM hidden layer sizes of ${10, 40, 70}$. For PLM-AS, the LSTM hidden layer size is fixed at 1024 to match the LM's dimensionality in \citet{Yang2023PLMASPL}.

    \item For \textbf{Eyettention}, we also train with a learning rate of 0.001 and dropout of 0.2, as done in \citep{Deng_Eyettention2023}

    \item For \textbf{PostFusion-Eye}, the 1D convolution layers have a kernel size of three, stride 1 and padding 1.

\end{itemize}

We train the deep-learning based models for a maximum of 40 epochs, with early stopping after 8 epochs if no improvement in the validation error is observed. Following \citet{liu_roberta_2019,mosbach2021on,Shubi2024Finegrained} we use the AdamW optimizer \cite{loshchilov_decoupled_2018} with a batch size of $16$. MAG-Eye, RoBERTa-Eye and PostFusion-Eye use a linear warm-up ratio of $0.06$, and a weight decay of 0.1. We standardize each eye movements feature using statistics computed on the training set, to zero mean unit variance. 

The code base for this project was developed with the assistance of GitHub Copilot, an AI-powered coding assistant. All generated code was carefully reviewed.

\clearpage
\section{Additional Results}
\label{app-sec:additional-res}

Below we present the test set ROC curves across
the ten cross-validation splits and their corresponding AUROC scores
(mean and standard deviation). We also provide accuracy results for the validation set.

\begin{figure*}[ht!]
    \centering
    \includegraphics[width=\textwidth]{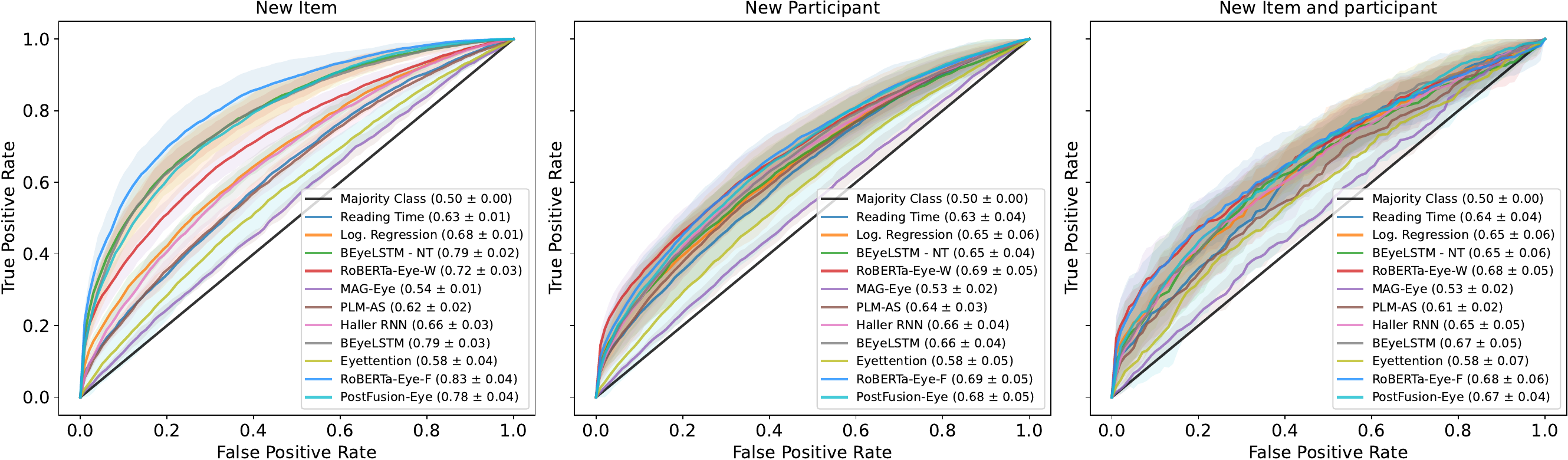}
    \caption{ROC Curves by model and evaluation regime. Each curve represents a different model across the ten cross-validation splits, with the corresponding AUROC scores (mean and standard deviation) provided in the legend.}
    \label{app-fig:roc_curves}
\end{figure*}

\begin{table*}[ht]
\centering
\resizebox{\textwidth}{!}{%
\begin{tabular}{@{}lccllll@{}}
\toprule
\textbf{Model} & {\makecell{\textbf{Gaze} \\ \textbf{Representation}}} & {\makecell{\textbf{Text} \\ \textbf{Representation}}} & {\makecell{\textbf{New} \\ \textbf{Item}}} & {\makecell{\textbf{New} \\ \textbf{Participant}}} & {\makecell{\textbf{New Item}\\ \& \textbf{Participant}}} & {\makecell{\textbf{All}}} \\
\midrule
Majority Class / Chance & -- & -- 
& ${50.0 _{\pm 0.0}}_{\text{\scriptsize +++}}$  
& ${50.0 _{\pm 0.0}}_{\text{\scriptsize +++}}$ 
& ${50.0 _{\pm 0.0}}_{\text{\scriptsize +++}}$ 
& ${50.0 _{\pm 0.0}}_{\text{\scriptsize +++}}$  \\

Reading Time & -- & -- 
& ${58.9 _{\pm 0.5}}_{\text{\scriptsize +++}}$ 
& ${58.9 _{\pm 1.0}}_{\text{\scriptsize +++}}$ 
& ${60.4 _{\pm 1.3}}_{\text{\scriptsize +++}}$ 
& ${58.9 _{\pm 0.5}}_{\text{\scriptsize +++}}$ \\ \addlinespace[1ex]
\hline
\addlinespace[1ex]
Log. Regression \citep{meziere2023using} & Global & -- 
& ${62.6 _{\pm 0.3}}_{\text{\scriptsize +++}}^{**}$ 
& ${60.6 _{\pm 1.5}}_{\text{\scriptsize +++}}$ 
& ${61.0 _{\pm 1.8}}_{\text{\scriptsize +++}}$ 
& ${61.6 _{\pm 0.8}}_{\text{\scriptsize +++}}^{*}$  \\

BEyeLSTM - No Text & Fixations & -- 
& ${73.2 _{\pm 0.6}}_{\text{\scriptsize +++}}^{***}$ 
& ${64.9 _{\pm 1.0}}_{\text{\scriptsize +++}}^{***}$ 
& ${65.1 _{\pm 1.4}}_{\text{\scriptsize +++}}^{*}$ 
& ${68.8 _{\pm 0.5}}_{\text{\scriptsize +++}}^{***}$ \\

ConvNext v2 & Image of Scanpath & -- 
& ${71.2 _{\pm 0.5}}_{\text{\scriptsize +++}}^{***}$ 
& ${65.3 _{\pm 0.9}}_{\text{\scriptsize +++}}^{***}$ 
& ${65.3 _{\pm 1.3}}_{\text{\scriptsize +++}}^{*}$ 
& ${68.0 _{\pm 0.5}}_{\text{\scriptsize +++}}^{***}$ \\

ViT & Image of Scanpath & -- 
& ${71.7 _{\pm 0.3}}_{\text{\scriptsize +++}}^{***}$ 
& ${65.8 _{\pm 0.9}}_{\text{\scriptsize +++}}^{***}$ 
& ${67.4 _{\pm 0.7}}_{\text{\scriptsize +++}}^{***}$ 
& ${68.6 _{\pm 0.5}}_{\text{\scriptsize +++}}^{***}$ \\

\addlinespace[1ex]
\hline
\addlinespace[1ex]
RoBERTa-Eye-W \citep{Shubi2024Finegrained} & Words & Emb+LF
& ${65.1 _{\pm 0.6}}_{\text{\scriptsize +++}}^{***}$ 
& ${64.9 _{\pm 1.1}}_{\text{\scriptsize +++}}^{***}$ 
& ${65.2 _{\pm 1.4}}_{\text{\scriptsize +++}}^{*}$ 
& ${65.1 _{\pm 0.6}}_{\text{\scriptsize +++}}^{***}$ \\

MAG-Eye \citep{Shubi2024Finegrained} & Words & Emb+LF 
& ${53.7 _{\pm 0.2}}_{\text{\scriptsize +++}}$ 
& ${53.5 _{\pm 0.5}}_{\text{\scriptsize +++}}$ 
& ${52.6 _{\pm 0.6}}_{\text{\scriptsize +++}}$ 
& ${53.5 _{\pm 0.2}}_{\text{\scriptsize +++}}$ \\
\addlinespace[1ex]
\hline
\addlinespace[1ex]
PLM-AS \citep{Yang2023PLMASPL} & Fixations Order & Emb 
& ${59.1 _{\pm 0.5}}_{\text{\scriptsize +++}}$ 
& ${61.1 _{\pm 0.7}}_{\text{\scriptsize +++}}$ 
& ${58.6 _{\pm 0.9}}_{\text{\scriptsize +++}}$ 
& ${60.1 _{\pm 0.4}}_{\text{\scriptsize +++}}$ \\

Haller RNN \citep{haller2022eye} & Fixations & Emb 
& ${62.3 _{\pm 0.6}}_{\text{\scriptsize +++}}^{**}$ 
& ${62.9 _{\pm 1.2}}_{\text{\scriptsize +++}}^{**}$ 
& ${63.4 _{\pm 1.3}}_{\text{\scriptsize +++}}$ 
& ${62.5 _{\pm 0.6}}_{\text{\scriptsize +++}}^{**}$ \\

BEyeLSTM \citep{reich2022inferring} & Fixations & LF 
& ${72.3 _{\pm 0.6}}_{\text{\scriptsize +++}}^{***}$ 
& ${65.0 _{\pm 1.3}}_{\text{\scriptsize +++}}^{***}$ 
& ${66.1 _{\pm 1.2}}_{\text{\scriptsize +++}}^{**}$ 
& ${68.5 _{\pm 0.6}}_{\text{\scriptsize +++}}^{***}$ \\

Eyettention \citep{Deng_Eyettention2023} & Fixations & Emb+LF 
& ${56.4 _{\pm 0.8}}_{\text{\scriptsize +++}}$ 
& ${56.6 _{\pm 0.9}}_{\text{\scriptsize +++}}$ 
& ${58.6 _{\pm 1.1}}_{\text{\scriptsize +++}}$ 
& ${56.6 _{\pm 0.5}}_{\text{\scriptsize +++}}$ \\

RoBERTa-Eye-F \citep{Shubi2024Finegrained} & Fixations & Emb+LF 
& ${90.7 _{\pm 0.3}}^{***}$ 
& ${91.9 _{\pm 0.5}}^{***}$ 
& ${88.7 _{\pm 0.9}}^{***}$ 
& ${91.2 _{\pm 0.3}}^{***}$ \\

PostFusion-Eye \citep{Shubi2024Finegrained} & Fixations & Emb+LF 
& ${89.2 _{\pm 0.4}}_{\text{\scriptsize +++}}^{***}$ 
& ${91.3 _{\pm 0.5}}^{***}$ 
& ${87.8 _{\pm 0.6}}^{***}$ 
& ${90.1 _{\pm 0.4}}_{\text{\scriptsize +++}}^{***}$ \\
\addlinespace[1ex]
\hline
\addlinespace[1ex]
Logistic Ensemble &  & 
& ${92.3 _{\pm 0.9}}^{***}$ 
& ${93.2 _{\pm 1.2}}^{***}$ 
& ${89.6 _{\pm 2.3}}^{***}$ 
& ${92.6 _{\pm 0.8}}^{***}$ \\
\bottomrule
\end{tabular}%
}
\small
\caption{Validation accuracy results aggregated across 10 cross-validation splits. `Emb' stands for word embeddings, `LF' for linguistic word features such as word length, frequency and surprisal, and `Fix' for fixations. Model performance is compared to the Reading Time baseline using a linear mixed effects model. In R notation: $is\_correct \sim model + (model \mid participant) + (model \mid paragraph)$. Significant gains over this baseline are marked with '*' $p < 0.05$, '**' $p < 0.01$ and '***' $p < 0.001$ in superscript, and significant drops compared to the best model in each regime are marked in subscript with '+'.}
\label{tab:eval-results}
\end{table*}

\clearpage
\section{Pairwise agreement between models by evaluation regime}
\label{app-sec:model_correlation_by_regime}

\begin{figure}[ht]
    \centering
    \includegraphics[width=1\linewidth]{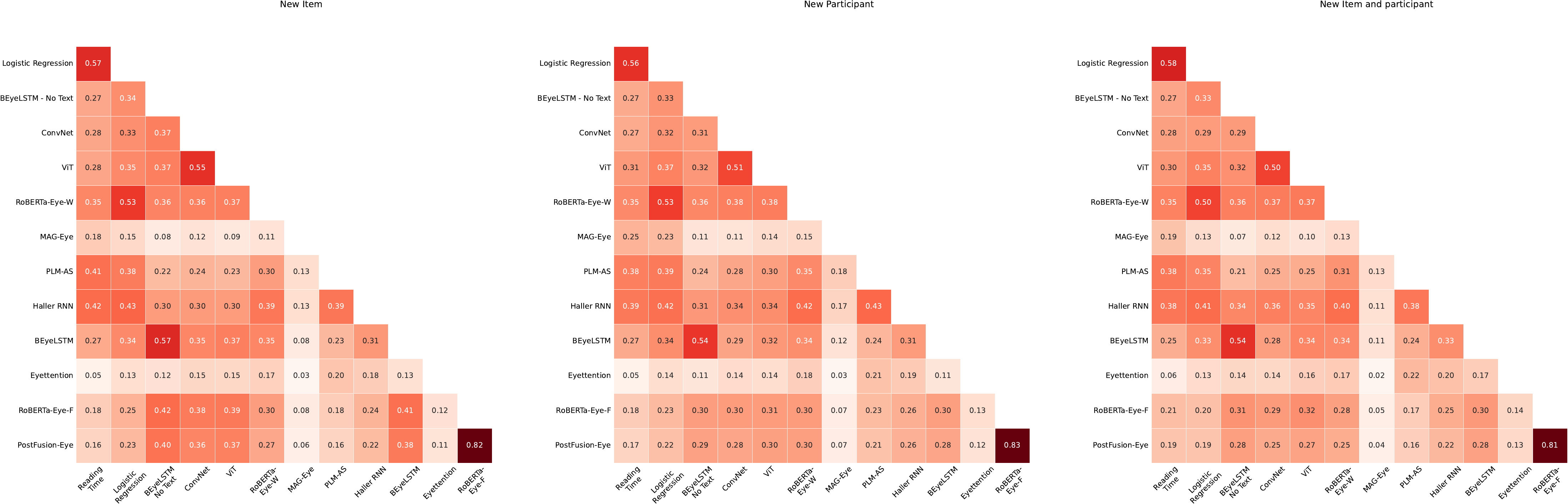}
    \caption{Pairwise Cohen's Kappa agreement between model predictions on the validation set by evaluation regime.}
    \label{app-fig:model_correlation_by_regime}
\end{figure}

\section{Feature Descriptions}
\label{sec:app-features}

We define 10 features that capture various aspects of the trial. These include the following reader features over the item:

\begin{enumerate}

    \item[1-3.] \textbf{Reading time before, within, and after critical span}: These features are motivated by the findings of \citet{malmaud_bridging_2020} and \cite{shubi2023cogsci} regarding faster reading times in information seeking compared to ordinary reading, primarily before and after the critical span, as well as the reported classification results of the Reading Time baseline.
    
    \addtocounter{enumi}{3}
    \item \textbf{Paragraph position} (1-54): Each participant reads 54 paragraphs, in a random article order. This feature captures the position of the paragraph in the experiment's presentation sequence. It is included as reading strategies can change as the experiment progresses (e.g. \citet{meiri2024dejavu} show that readers become faster as the experiment progresses).
    
    \item \textbf{Answered correctly}: this feature encodes whether after having read the passage, the participant answered the given reading comprehension question correctly. It captures participant-specific task difficulty and the extent to which the participant read the passage attentively.
    
\end{enumerate}
We further include the following item (paragraph and question), reader-independent features: 
\begin{enumerate}
  \setcounter{enumi}{5}
    \item \textbf{Paragraph length} (in words): this feature is chosen as we hypothesize that more data could lead to more accurate predictions for the item.
    \item \textbf{Paragraph level} (Advanced / Elementary): is chosen as eye movements could be influenced by the difficulty level of the text, for example through differences in word frequency and surprisal \citep{singh_etal_2016_quantifying,hollenstein_etal_2022_patterns}. 
    
    \item \textbf{Critical span start location} (relative position, normalized by paragraph length): \citet{shubi2023cogsci} showed skimming-like reading patterns after processing task critical information in information seeking. We thus hypothesize that earlier appearance of task critical information could facilitate the ability to correctly identify information seeking reading.
    \item \textbf{Critical span length} (normalized by paragraph length): we also hypothesize that less task critical information during information seeking could further aid distinguishing it from ordinary reading. 
    
    \item \textbf{Question difficulty} (percentage of participants who answered the question incorrectly): estimated from the train data. We include this feature as it can influence eye movements in information seeking, with harder questions potentially obscuring patterns of goal oriented reading.
    
\end{enumerate}

\end{document}